\documentclass{article}

\PassOptionsToPackage{numbers,sort&compress}{natbib}
\usepackage[preprint]{neurips_2026}

\usepackage[utf8]{inputenc}
\usepackage[T1]{fontenc}
\usepackage{amsmath}
\usepackage{amssymb}
\usepackage{amsfonts}
\usepackage{booktabs}
\usepackage{graphicx}
\usepackage{algorithm}
\usepackage{algorithmic}
\usepackage{multirow}
\usepackage{makecell}
\usepackage{float}
\usepackage{tabularx}
\usepackage{array}
\usepackage[table]{xcolor}
\usepackage{caption}
\usepackage{nicefrac}
\usepackage{microtype}
\usepackage{url}
\usepackage{hyperref}
\usepackage{tcolorbox}
\usepackage{tikz}
\usepackage{hyperref}
\usepackage{booktabs}
\usepackage{multirow}
\usepackage{makecell}
\usepackage[table]{xcolor}

\definecolor{lightyellow}{RGB}{255,245,157}
\definecolor{mygreen}{RGB}{0,128,0}
\definecolor{myred}{RGB}{200,0,0}

\definecolor{rowgray}{RGB}{242,242,242}
\definecolor{mygreen}{RGB}{96,160,92}
\definecolor{myorange}{RGB}{220,140,60}

\definecolor{rowgray}{RGB}{245,245,245}
\definecolor{mygreen}{RGB}{0,150,0}
\definecolor{mypurple}{RGB}{150,0,150}
\definecolor{deltagreen}{RGB}{96,160,92}
\newcommand{\gup}[1]{\textcolor{deltagreen}{\scriptsize$\uparrow$#1}}
\newcommand{\gdown}[1]{\textcolor{deltagreen}{\scriptsize~$\downarrow$#1}}
\newcommand{\Name}{UniShield}

\let\mathDelta\Delta
\renewcommand{\Delta}{\ensuremath{\mathDelta}}
\begin{document}

\title{\Name: Unified Face Attack Detection via KG-Informed Multimodal Reasoning} 

\author{%
\normalfont\small
Hongrui Li$^{1,*}$, Yichen Shi$^{2,*}$, Hongyang Wang$^{1}$, Yuhao Gao$^{3}$, Hui Ma$^{4}$, Jun Feng$^{1,\dagger}$, Zitong Yu$^{4,\dagger}$\\
\normalfont\small
$^{1}$Shijiazhuang Tiedao University \quad
$^{2}$Shanghai Jiao Tong University\\
\normalfont\small
$^{3}$Ningbo Institute of Digital Twin \quad
$^{4}$Great Bay University
}

\maketitle
\begingroup
\renewcommand{\thefootnote}{\fnsymbol{footnote}}
\footnotetext[1]{Equal contribution.}
\footnotetext[2]{Corresponding authors.}
\endgroup

\vspace{-0.8em}

\begin{abstract}
% Unified face attack detection (UAD) aims to address both physical and digital spoofing within a single framework. Existing methods, however, are limited by task-specific pipelines and fail to establish cross-type semantic alignment, hindering their ability to model relationships across attack modalities and limiting reliable generalization. To bridge this gap, we propose UniShield, a novel benchmark for unified face attack detection, and introduce a knowledge-graph-driven multimodal large language model (MLLM) framework. Our framework explicitly aligns semantics across heterogeneous spoofing types through a unified face attack knowledge graph (FAKG), which connects attack categories with observable visual cues and their underlying causes. We leverage FAKG to automatically generate high-quality question-answer pairs for Attack-Graph Instruction Tuning (AGIT) of a multimodal foundation model. Additionally, we present Graph-Consistent Reasoning Optimization (GCRO), a GRPO-based reasoning optimization objective that regularizes the model's reasoning trajectories to maintain consistency with the knowledge graph, improving the reliability and faithfulness of its predictions. Extensive experiments on UniShield and existing protocols demonstrate that our framework sets a new standard for unified face attack detection, achieving state-of-the-art performance with more reliable and interpretable predictions.

Unified face attack detection (UAD) requires recognizing physical spoofing and digital forgery within a shared decision space, yet existing discriminative or prompt-based methods largely rely on appearance correlations and provide limited evidence-grounded reasoning. We propose UniShield, a knowledge-grounded multimodal reasoning framework for unified face attack defense. UniShield constructs a Face Attack Knowledge Graph (FAKG) that links attack categories to diagnostic visual cues and attack-conditioned relations, and uses it to synthesize 52,025 FAKG-QA examples for Attack-Graph Instruction Tuning (AGIT). To improve rationale consistency, we further introduce Graph-Consistent Reasoning Optimization (GCRO), a GRPO-based objective with a KG-consistency reward that encourages generated rationales to match graph-supported cues while penalizing incompatible claims. Experiments on our multimodal UAD benchmark show that UniShield achieves strong performance across binary, coarse-grained, and fine-grained protocols, with consistently high ACC and low HTER. These results suggest that structured attack knowledge can improve both detection accuracy and reasoning reliability over discriminative baselines and general-purpose MLLMs. Our code will be released at \url{https://anonymous.4open.science/r/Unishield-A6A3/}.

% Unified Face Attack Detection seeks to identify heterogeneous spoofing attacks within a single unified framework, yet existing approaches remain limited to task-specific designs and lack semantic reasoning capability. In this paper, we present the first unified face attack defense framework that integrates multimodal large language models to enhance detection performance, robustness, and interpretability. We establish a new benchmark for unified face attack detection and propose a knowledge-driven supervised fine-tuning paradigm. Specifically, we construct a unified face attack knowledge graph that encodes attack types, visual cues, and underlying physical or digital factors, which is leveraged to automatically generate high-quality question–answer pairs for supervised fine-tuning of a multimodal foundation model. Furthermore, we incorporate conformal set calibration to transform point predictions into calibrated prediction sets, enabling uncertainty-aware inference. To refine ambiguous cases, general-purpose MLLMs are employed to generate attack descriptions and image captions for secondary reasoning and correction. Extensive experiments demonstrate that the proposed framework consistently outperforms existing methods while providing reliable and interpretable predictions.
  % \keywords{Unified face attack detection \and Multimodal large language models \and  Knowledge graph based learning}
\end{abstract}

\section{Introduction}
\label{intro}

Face attack detection plays a pivotal role in modern biometric security systems, safeguarding applications such as mobile authentication, access control, and digital identity verification. 
With the rapid evolution of attack techniques—from traditional print and replay spoofing to sophisticated face forgery and video-driven manipulation—the boundary between physical and digital attacks has become increasingly blurred. 
Recent research has therefore shifted toward Unified face attack detection (UAD)~\cite{uni,uni++}, which aims to handle heterogeneous attack types within a single framework. 
Such a unified paradigm is essential for real-world deployment, where systems must generalize across diverse attack forms while maintaining robustness, interpretability, and scalability.
    
Despite recent progress, most existing UAD methods still treat unified face attack detection primarily as a label prediction problem. 
As summarized in Table~\ref{tab:method_categorization}, traditional deep learning methods learn visual decision boundaries with CNN or Vision Transformer backbones; prompt-learning methods adapt vision--language models through fixed or learnable prompts; and recent MLLM-based methods introduce textual explanations for face-security decisions. 
However, these paradigms lack an explicit mechanism to connect attack categories, diagnostic visual evidence, and reasoning rationales. 
As a result, their predictions and explanations can rely on superficial correlations rather than attack-specific forensic knowledge, limiting reliability under heterogeneous physical and digital attacks.

To address these challenges, we propose \textbf{UniShield}, a unified MLLM-based face attack defense framework. 
Unlike conventional parallel pipelines, UniShield adopts a single-stream reasoning paradigm that tightly integrates visual encoding, language modeling, and structured knowledge guidance. 
At the core of our framework lies a unified Face Attack Knowledge Graph (FAKG), which explicitly models attack types, shared artifacts, and cross-type semantic relations. 
By injecting FAKG into multimodal large language model reasoning, UniShield enables knowledge-aware inference trajectories rather than free-form generation. 
This design supports multi-level outputs within a single forward pass, including: 
(1) basic real/attack judgment, 
(2) fine-grained attack type classification across physical and digital categories, and 
(3) interpretable reasoning rationales grounded in visual cues and structured semantic evidence. 
Through knowledge-guided supervised fine-tuning and reinforcement learning optimization, UniShield improves unified face attack detection with more interpretable and semantically grounded predictions.

Extensive experiments on a newly constructed unified benchmark demonstrate that UniShield consistently outperforms existing traditional, prompt-based, and MLLM-based baselines, while providing fine-grained predictions and interpretable reasoning. 
These results highlight the importance of moving beyond appearance-level discrimination toward structured, knowledge-guided multimodal reasoning for unified face attack defense.

% \begin{figure}[h]
%   \centering
%   \includegraphics[width=0.98\linewidth]{figures/newintro.png}
%   \caption{Conceptual comparison between discriminative UAD and UniShield. (a) Conventional methods mainly perform coarse real/attack classification from visual features or prompts. (b) UniShield injects FAKG-guided multimodal reasoning to produce judgment, fine-grained attack type, and interpretable reasoning cues.}
%   \label{fig:previous}
%   \vspace{-5mm}
% \end{figure}

\renewcommand\tabularxcolumn[1]{m{#1}}
\newcolumntype{Y}{>{\centering\arraybackslash}X}
\newcolumntype{L}{>{\raggedright\arraybackslash}X}

\begin{table}[t]
\centering
\scriptsize
\setlength{\tabcolsep}{2pt}
\renewcommand{\arraystretch}{0.96}

\caption{Categorization of existing methods for Face Anti-Spoofing, Face Forgery Detection, and Unified Face Attack Detection under different learning paradigms.}

\begin{tabularx}{\linewidth}{>{\centering\arraybackslash}m{1.85cm} L L L}
\toprule
\textbf{Method Type} & \multicolumn{1}{c}{\textbf{FAS}} & \multicolumn{1}{c}{\textbf{FFD}} & \multicolumn{1}{c}{\textbf{UAD}} \\
\midrule

\textbf{Traditional Deep Learning} &
AS-FAS~\cite{Liu2018}, 
CDCN~\cite{cdcn}, 
Yang et~al.~\cite{yang2014},
Atoum et~al.~\cite{atoum2017} &
Face X-ray~\cite{facexraytradffd},
Capsule-Forensics~\cite{capsuletrffd},
Rossler et~al.~\cite{rossler2019}, 
Guarnera et~al.~\cite{guarnera2020deepfake} &
Yu et al.~\cite{yujoint} \\

\midrule

\textbf{Prompt Learning} &
FGPL~\cite{liufasprompt}, 
Cfpl-fas~\cite{hufasprompt} &
DeepfakeCLIP~\cite{deepfakeclipffd}, 
C2pclip~\cite{tanffdprompt}, 
KGPL~\cite{wangffdprompt} &
La-SoftMoE CLIP~\cite{la-clip}, 
UniAttackDetection~\cite{uni}, 
HiPTune~\cite{uni++} \\

\midrule

\textbf{MLLM} &
\mbox{FaceShield~\cite{wangfaceshield}}, 
\mbox{CEPL~\cite{zhangfasmllm}} &
\mbox{LVLM-DFD~\cite{yuffdmllm}},
\mbox{VLForgery~\cite{heffdmllm}} &
\mbox{Ours} \\

\bottomrule
\end{tabularx}

\label{tab:method_categorization}
\vspace{-1mm}
\end{table}

Our main contributions are summarized as follows:

\begin{itemize}

\item \textbf{A Reasoning-based Unified Face Attack Defense Framework.}
% We propose UniShield, the first framework to transition UAD from traditional discriminative pattern matching to reasoning-based cognitive vision. By leveraging Multimodal Large Language Models (MLLMs), UniShield provides a single-stream, interpretable solution that unifies physical spoofing and digital forgery detection under a cohesive reasoning logic.
We propose UniShield, a unified MLLM-based framework that shifts UAD from pattern matching to knowledge-guided multimodal reasoning. UniShield provides a single-stream, interpretable solution for detecting both physical spoofing and digital forgery attacks.

% \item \textbf{Expert-Level Knowledge Distillation via FAKG.}
% We introduce the FAKG, a structured ‘cognitive map’ that systematically codifies domain expertise into a unified semantic space. FAKG serves as the first explicit knowledge foundation for UAD, bridging the gap between low-level visual artifacts and high-level attack causality across heterogeneous spoofing modalities.

\item \textbf{FAKG-Guided Knowledge Injection and Instruction Tuning.}
We construct a Face Attack Knowledge Graph (FAKG) that organizes attack types, diagnostic cues, and cross-type semantic relations into a unified reasoning space. Based on FAKG, we develop Attack-Graph Instruction Tuning (AGIT), which converts attack-cue subgraphs into multi-hop reasoning Q\&A pairs for supervised fine-tuning, enabling structured knowledge injection and cross-attack generalization.

\item \textbf{Graph-Consistent Reasoning Optimization for Faithful Reasoning.}
% We propose GCRO, a specialized reinforcement learning algorithm designed to enforce logical faithfulness. By explicitly penalizing deviations between the model's internal thinking trajectories and the FAKG’s causal structure, GCRO ensures that the model's predictions are not just statistically accurate but are KG-consistent, effectively mitigating logical hallucinations in security-critical scenarios.
We propose Graph-Consistent Reasoning Optimization (GCRO), a GRPO-based objective that optimizes predictions with FAKG-consistency supervision. By encouraging rationales to match graph-supported cues and penalizing graph-inconsistent claims, GCRO promotes more KG-consistent reasoning in security-critical scenarios.

\end{itemize}

% \begin{itemize}

% \item \textbf{UniShield: The First Unified MLLM-based Face Attack Defense Framework.}
% We propose UniShield, the first unified framework that integrates multimodal large language models into Unified Face Attack Detection, enabling a single-stream solution for both Face Spoofing and Forgery Detection with interpretable reasoning.

% \item \textbf{Construction of a Unified Face Attack Knowledge Graph (FAKG).}
% We design a structured knowledge graph that systematically models attack categories, shared artifacts, and cross-type semantic relationships, providing an explicit reasoning foundation for unified detection.

% \item \textbf{FAKG-Guided High-Quality QA Dataset Construction.}
% Based on the proposed FAKG, we automatically generate a large-scale, structured question–answer dataset for supervised fine-tuning, enabling effective knowledge injection and cross-attack generalization.

% \item \textbf{Knowledge-Consistent Reinforcement Learning with Think–FAKG Alignment.}
% We introduce a reinforcement learning strategy where the reward explicitly measures consistency between intermediate reasoning (``think'') trajectories and FAKG structure, encouraging semantically grounded and logically coherent predictions.

\section{Related Work}
\label{relatedwork}
\subsection{Deep and Prompt Learning for Face Attack Detection}
Face attack detection has long been dominated by task-specific discriminative modeling.
For FAS, CNN-based methods model liveness cues such as texture, frequency, or physiological signals~\cite{Liu2018,cdcn,atoum2017,yang2014}; for FFD, detectors focus on manipulation artifacts and spatial inconsistencies~\cite{facexraytradffd,capsuletrffd,rossler2019,guarnera2020deepfake}.
Early UAD efforts extend this formulation to mixed physical--digital attacks~\cite{yujoint}, but still mainly rely on appearance-level decision boundaries that are sensitive to dataset-specific artifacts and domain shifts.
% Recent studies therefore explore vision-language model (VLM) adaptation to improve transferability by aligning visual evidence with language semantics.
% A line of work leverages prompt learning to adapt CLIP-like VLMs for security tasks: FGPL~\cite{liufasprompt} and Cfpl-fas~\cite{hufasprompt} for FAS, and DeepfakeCLIP~\cite{deepfakeclipffd}, C2PCLIP~\cite{tanffdprompt}, and KGPL~\cite{wangffdprompt} for FFD.
% Beyond single-task settings, unified detection is formulated to handle both physical and digital attacks within a shared framework~\cite{uni}.
% To further mitigate modality/domain gaps, La-SoftMoE CLIP~\cite{la-clip} introduces soft mixture-of-experts to improve adaptability, while HiPTune~\cite{uni++} structures prompts to encode multi-level attack semantics for UAD.
% Overall, VLM-based methods enhance cross-modal alignment and domain robustness. However, they remain primarily discriminative and depend on fixed or semi-fixed prompt templates, which limits structured reasoning, fine-grained interpretability, and controllable knowledge utilization in unified scenarios.
Recent VLM and prompt-learning methods improve transferability by adapting CLIP-like models to FAS and FFD~\cite{liufasprompt,hufasprompt,deepfakeclipffd,tanffdprompt,wangffdprompt}, and to UAD through shared label spaces, soft mixture-of-experts, and hierarchical prompts~\cite{uni,la-clip,uni++}.
However, they still primarily produce label predictions through fixed or learned prompts, without explicitly grounding decisions in structured attack-cue relations.

\subsection{MLLMs for Multimodal Reasoning in Face Security}
MLLMs align a vision encoder with large language models, enabling joint visual understanding and textual generation with emergent reasoning capabilities.
General alignment paradigms such as BLIP-2~\cite{blip2} and MiniGPT-4~\cite{minigpt4} demonstrate strong instruction-following and multimodal reasoning, motivating their adoption in security-oriented perception.

In face security, emerging MLLM-based approaches explore explainable detection.
For FAS, FaceShield~\cite{wangfaceshield} employs MLLMs to provide spoofing decisions with textual explanations, and CEPL~\cite{zhangfasmllm} further investigates promptable/consistent evidence learning.
For FFD, LVLM-DFD~\cite{yuffdmllm} and VLForgery~\cite{heffdmllm} extend MLLMs to detect and describe forgery traces with language grounding.
In addition, benchmarks such as SHIELD~\cite{shield} study MLLM behaviors under spoofing and forgery settings, indicating both opportunities and vulnerabilities.
Despite improved reasoning and explainability, existing MLLM-based methods typically rely on free-form generation without explicitly modeling structured attack knowledge, which can reduce robustness and fine-grained controllability for unified physical-digital detection.
To address this limitation, we introduce UniShield, which integrates structured FAKG reasoning into MLLM inference, enabling attack-aware multimodal reasoning guided by explicit relational knowledge and improving reliability in unified detection.

\section{Methods}
\label{methods}
\subsection{Knowledge-Driven Cognitive Foundation: The FAKG}
\label{dataset gen}

\begin{figure}[t]
  \centering
  \includegraphics[width=0.85\linewidth]{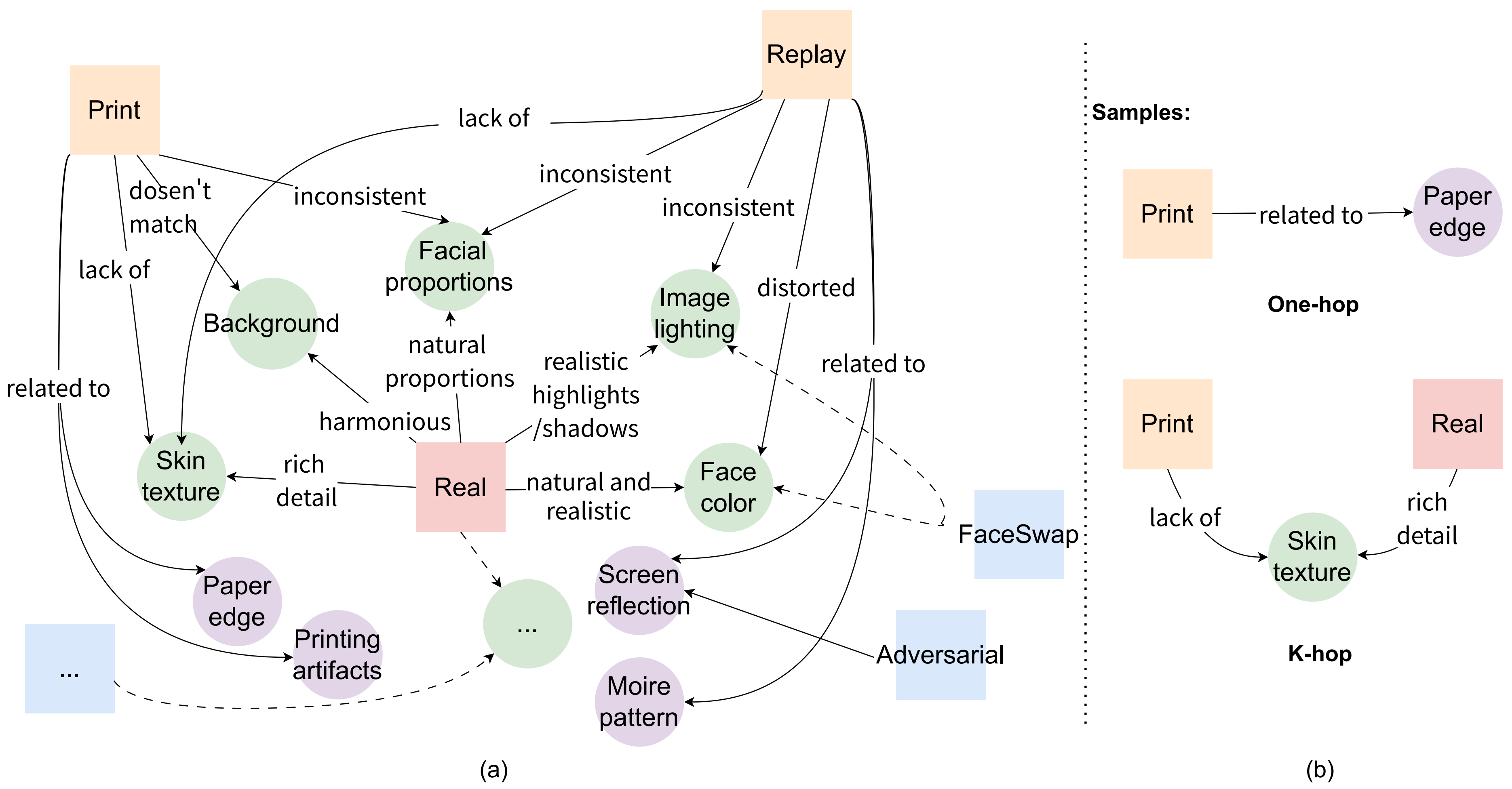}
  \caption{Face Attack Knowledge Graph. Squares denote attack types (color-coded); circles denote features. \textcolor{mygreen}{Green} circles are shared features across attacks, and \textcolor{mypurple}{purple} circles are type-specific features; edges indicate relations.}
  \label{fig:fakg}
  \vspace{-1mm}
\end{figure}

\begin{figure}[t]
  \centering
  \includegraphics[width=\linewidth]{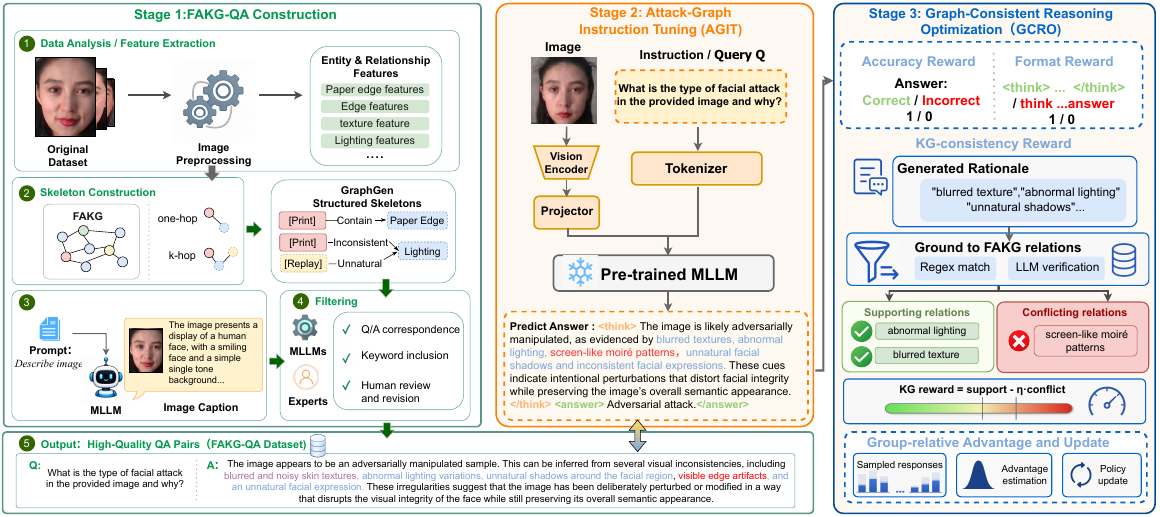}
  \caption{Overview of the UniShield framework, including FAKG-guided QA dataset construction, AGIT, and GCRO for graph-consistent multimodal reasoning in UAD.}
  \label{fig:framework}
  \vspace{-1mm}
\end{figure}

To transcend the limitations of label-centric discriminative models, we transition towards a reasoning-based paradigm by constructing the FAKG. Unlike traditional flat labeling, FAKG organizes domain-informed attack-cue relations between heterogeneous attack modalities and their observable physical/digital manifestations, as illustrated in Fig.~\ref{fig:fakg}. This graph functions as a relation-aware semantic scaffold, providing a unified foundation that bridges the gap between low-level visual perception and high-level forensic reasoning~\cite{uni,uni++}. The overall UniShield framework built on this graph is summarized in Fig.~\ref{fig:framework}.

We formalize the FAKG as a heterogeneous semantic graph:\begin{equation}\mathcal{G} = (\mathcal{E}, \mathcal{R}),\end{equation}where $\mathcal{E}$ denotes the set of cognitive entities and $\mathcal{R}$ represents the directed attack-cue relations captured by the graph. Specifically, $\mathcal{E}$ comprises a dual-layered hierarchy:

\textbf{Attack-type Entities} define the taxonomy of the defense space, encompassing Bona Fide faces and a spectrum of threats across the physical-digital divide (e.g., Print, Replay, Adversarial, FaceSwap).

\textbf{Feature Entities} represent the observable diagnostic cues, categorized into Inherent Common Features (e.g. skin texture, global illumination) and Attack-Specific Artifacts (e.g. lattice patterns, geometric inconsistencies). This bifurcation allows the model to perform joint-distribution modeling of shared regularities and idiosyncratic anomalies.

The relations within FAKG encode directed semantic dependencies and are defined as triples:\begin{equation}(a,\rho,f)\in\mathcal{R},\end{equation}where $a$ is the source attack entity, $f$ is the target feature, and $\rho$ denotes the attack-conditioned relation describing how $f$ manifests under $a$.

\textbf{This relation-aware formulation is useful for disambiguation:} it helps the system distinguish visually similar artifacts by contextualizing them with attack-specific forensic cues. Consequently, FAKG reformulates UAD as a knowledge-informed diagnostic process and encourages the model's generated rationales to remain connected to established forensic cues.

\subsection{FAKG-Guided Dataset Synthesis}
To construct a high-quality and reasoning-oriented QA dataset, we propose an FAKG-guided synthesis pipeline that integrates structured graph reasoning with multimodal captioning, as shown in Fig.~\ref{fig:framework}.
Given an annotated face anti-spoofing dataset $\mathcal{D}=\{(x_i, y_i)\}_{i=1}^{N}$ (image $x_i$ with attack label $y_i$) and the Face Attack Knowledge Graph $\mathcal{G}=(\mathcal{E},\mathcal{R})$, our goal is to synthesize a QA corpus
\begin{equation}
\mathcal{Q}^{\star}=\{(x_i, q_i, a_i)\}_{i=1}^{M},
\end{equation}
where each QA pair is grounded in both structured KG knowledge and image-specific evidence.

% \textbf{Stage 1: KG-Driven QA Template Generation.}
% For each label $y$, we first extract a local subgraph (or path set) centered at the corresponding attack node $e_y\!\in\!\mathcal{E}$.
% The $1$-hop neighborhood subgraph is defined as
% \begin{equation}
% \mathcal{S}^{(1)}(y)=\{(e_y, r, e)\mid (e_y,r,e)\in\mathcal{R}\},
% \end{equation}
% which induces single-relation reasoning templates (e.g., \emph{``Why is this a print attack?''}).
% To enable contrastive reasoning across attack types, we further sample a $k$-hop path set
% \begin{equation}
% \mathcal{S}^{(k)}(y)=\{(e_0,r_1,e_1,\ldots,r_k,e_k)\mid e_0=e_y,\ (e_{t-1},r_t,e_t)\in\mathcal{R}\},
% \end{equation}
% where $k>1$ allows the template to involve shared vs.\ attack-specific cues and supports questions such as
% \emph{``Why is this a print attack and not a replay attack?''}.
% We then feed the extracted subgraph into GraphGen~\cite{graphgen} to obtain a structured reasoning template:
% \begin{equation}
% \tau_i=\mathrm{GraphGen}(\mathcal{S}^{(k)}(y_i)).
% \end{equation}
% This process ensures that the generated templates are semantically grounded in FAKG.

\textbf{Stage 1: KG-Driven QA Skeleton Generation.}
For each sample label $y_i$, we locate its corresponding attack node $e_{y_i}\in\mathcal{E}$ in the knowledge graph $\mathcal{G}=(\mathcal{E},\mathcal{R})$.
We then extract a local $k$-hop ego-subgraph around $e_{y_i}$ as the input to GraphGen~\cite{graphgen}, so that the generated QA skeletons are grounded in the structured relations of FAKG.

\textbf{ $k$-hop Ego-subgraph Extraction.}
Let $\mathrm{dist}_{\mathcal{G}}(u, v)$ denote the shortest-path distance between nodes $u$ and $v$ in $\mathcal{G}$.
We define the $k$-hop neighborhood node set centered at $e_{y_i}$ as
\begin{equation}
\mathcal{E}^{(k)}(y_i)=\{e\in\mathcal{E}\mid \mathrm{dist}_{\mathcal{G}}(e, e_{y_i})\le k\},
\end{equation}
and take the induced edge set restricted to these nodes:
\begin{equation}
\mathcal{R}^{(k)}(y_i)=\{(u,\rho,v)\in\mathcal{R}\mid u\in\mathcal{E}^{(k)}(y_i),\ v\in\mathcal{E}^{(k)}(y_i)\}.
\end{equation}
The resulting local subgraph is
\begin{equation}
\mathcal{S}^{(k)}(y_i)=\big(\mathcal{E}^{(k)}(y_i),\mathcal{R}^{(k)}(y_i)\big).
\end{equation}
When $k{=}1$, $\mathcal{S}^{(1)}(y_i)$ mainly captures direct relations between an attack type and its associated cues, yielding single-relation skeletons such
 as \emph{``Why is this a print attack?''}.
When $k{>}1$, the subgraph additionally contains intermediate entities that connect shared cues and attack-specific cues, enabling multi-step and contrastive skeletons across attack types.

\textbf{Subgraph-to-skeleton Generation.}
We feed $\mathcal{S}^{(k)}(y_i)$ into GraphGen~\cite{graphgen} to produce a structured reasoning skeleton:
\begin{equation}
\tau_i=\mathrm{GraphGen}(\mathcal{S}^{(k)}(y_i)).
\end{equation}
In particular, multi-hop subgraphs allow GraphGen to incorporate both common and distinctive cues, resulting in contrastive questions such as
\emph{``Why is this a print attack and not a replay attack?''}.

\textbf{Stage 2: Multimodal Caption Generation.}
Given $(x_i,y_i)$, we employ an MLLM captioner to produce an evidence-focused description that highlights observable cues relevant to the label:
\begin{equation}
c_i=\mathrm{MLLM}_{\text{cap}}(x_i, y_i),
\end{equation}
where $c_i$ summarizes visual evidence such as illumination, texture consistency, boundary artifacts, and other KG-related features.

% \textbf{Stage 3: KG-Consistency Filtering.}
% To retain only semantically aligned and structurally valid samples, we apply a dual-level filter with an automatic verifier followed by expert review.
% We define an automatic consistency score $s_i\in[0,1]$ as a weighted combination of (i) \emph{keyword/evidence coverage} and (ii) \emph{graph-relation consistency}:
% \begin{equation}
% s_i=\alpha \cdot \mathrm{Cov}(c_i,\tau_i)\;+\;(1-\alpha)\cdot \mathrm{Consis}_{\mathcal{G}}(a_i,\tau_i),
% \end{equation}
% where $\mathrm{Cov}(c_i,\tau_i)$ measures whether the caption contains the key KG cues required by the template, and
% $\mathrm{Consis}_{\mathcal{G}}(a_i,\tau_i)$ evaluates whether the generated answer obeys the relations specified by $\tau_i$ (e.g., required cues present, contradictory cues absent).
% We keep samples that satisfy both semantic and format constraints:
% \begin{equation}
% \mathbb{I}_i=\mathbf{1}\big[s_i\ge\delta\ \wedge\ \texttt{format}(\tau_i)=\texttt{valid}\big],
% \end{equation}
% and further apply a human filter $\mathbb{H}_i\in\{0,1\}$ to remove ambiguous or noisy cases, yielding the retained set
% \begin{equation}
% \mathcal{D}^{\star}=\{(x_i,y_i)\mid \mathbb{I}_i=1\ \wedge\ \mathbb{H}_i=1\}.
% \end{equation}

\textbf{Stage 3: Knowledge--Caption Fusion for QA Synthesis.}
Finally, we fuse the KG-grounded skeleton and the image caption to synthesize the final QA pair through semantic fusion:
\begin{equation}
(q_i,a_i)=\mathrm{MLLM}_{\text{fus}}(x_i,\tau_i,c_i),\qquad (x_i,y_i)\in\mathcal{D}^{\star}.
\end{equation}
The resulting high-quality corpus is
\begin{equation}
\mathcal{Q}^{\star}=\{(x_i,q_i,a_i)\mid (x_i,y_i)\in\mathcal{D}^{\star}\},
\end{equation}
which integrates structured domain knowledge from $\mathcal{G}$ with instance-specific visual evidence, and provides discriminative and interpretable supervision for knowledge-driven training.

\subsection{Attack-Graph Instruction Tuning}
We adopt a two-stage training strategy. In the cold-start stage, we introduce Attack-Graph Instruction Tuning (AGIT), which converts FAKG-centered attack-cue subgraphs into evidence-grounded diagnostic instruction traces. In implementation, AGIT fine-tunes a pretrained multimodal large language model on our FAKG-guided QA dataset, so that the model learns to follow the required output format and produces a reasonable initial reasoning/prediction before reinforcement learning.
Each training sample is organized as:
\begin{equation}
(x, q, y^{\text{think}}, y^{\text{ans}}),
\end{equation}
where $x$ denotes the face image, $q$ is the knowledge-driven question, $y^{\text{think}}$ is the rationale, and $y^{\text{ans}}$ is the final prediction (attack label). We concatenate $y=[y^{\text{think}}, y^{\text{ans}}]$ and optimize the standard teacher-forcing objective:
\begin{equation}
\mathcal{L}_{\text{AGIT}}
=
-\sum_{t}\log P_{\theta}(y_t \mid x, q, y_{<t}).
\end{equation}

While AGIT provides a strong initialization through the standard SFT objective, it does not explicitly constrain the rationale to be consistent with the relations in FAKG, motivating the subsequent graph-consistent reasoning refinement.

\subsection{Graph-Consistent Reasoning Optimization}
% Following Group Relative Policy Optimization (GRPO)~\cite{guo2025deepseek}, we refine the model with rule-based rewards (accuracy and format) and our proposed KG-consistency reward, similar in spirit to rule-reward designs used for reasoning models.
To reduce unsupported or inconsistent rationales, we propose Graph-Consistent Reasoning Optimization (GCRO), a GRPO-based reasoning optimization objective with accuracy, format, and KG-consistency rewards. Built upon the Group Relative Policy Optimization (GRPO) framework~\cite{guo2025deepseek}, GCRO shifts the optimization focus from simple label accuracy to reasoning path consistency, encouraging the model's generated rationales to remain aligned with FAKG-supported cues.

Given an input $(x,q)$, GCRO samples a \emph{group} of $G$ responses:
\begin{equation}
y^{(g)}=\{y^{(g),\text{think}},y^{(g),\text{ans}}\}\sim \pi_{\theta_{\text{old}}}(\cdot \mid x,q), \quad g=1,\dots,G,
\end{equation}
and assigns each response a scalar reward:
\begin{equation}
R^{(g)}
=
\lambda_{\text{acc}}R^{(g)}_{\text{acc}}
+
\lambda_{\text{fmt}}R^{(g)}_{\text{fmt}}
+
\lambda_{\text{kg}}R^{(g)}_{\text{kg}}.
\end{equation}
Following the original GRPO setting~\cite{guo2025deepseek}, $R_{\text{acc}}$ and $R_{\text{fmt}}$ are implemented as rule-based answer-correctness and response-format checks; below we detail only our proposed KG-consistency reward.

\noindent\textbf{KG-consistency Reward.}
Our key contribution is a KG-consistency reward that encourages rationales to align with the domain relations encoded in FAKG. We decompose $R_{\text{kg}}$ into a positive matching reward and a contradiction penalty:
\begin{equation}
R_{\text{kg}}^{(g)}
=
\mathrm{clip}_{[0,1]}
\left(
R_{\text{match}}^{(g)}
-
\eta R_{\text{conflict}}^{(g)}
\right).
\end{equation}

To compute these terms, we project the generated reasoning text onto the existing FAKG relation space:
\begin{equation}
\begin{aligned}
\mathcal{C}^{(g)}
&=
\mathrm{Ground}\left(y^{(g),\text{think}},\mathcal{G}\right) \\
&=
\left\{
(a,\rho,f)\in\mathcal{R}
\ \middle|\
\mathrm{Match}_{\text{regex}}\left(y^{(g),\text{think}},a,\rho,f\right)
\lor
\mathrm{Match}_{\text{llm}}\left(y^{(g),\text{think}},a,\rho,f\right)
\right\}.
\end{aligned}
\end{equation}
Here, $\mathrm{Match}_{\text{regex}}$ denotes deterministic regular-expression matching with predefined patterns for each FAKG relation, and $\mathrm{Match}_{\text{llm}}$ verifies paraphrased or implicit mentions of the same relation. The LLM verifier is restricted to existing FAKG relations and cannot introduce new relations.

For the ground-truth attack type $a^\ast$, let $\mathcal{S}^{+}(a^\ast)$ denote the FAKG relations that support this attack type, and $\mathcal{S}^{-}(a^\ast)$ denote relations that are incompatible with it. We compute:
\begin{equation}
R_{\text{match}}^{(g)}
=
\frac{
|\mathcal{C}^{(g)}\cap \mathcal{S}^{+}(a^\ast)|
}{
|\mathcal{S}^{+}(a^\ast)|+\epsilon
},
\qquad
R_{\text{conflict}}^{(g)}
=
\frac{
|\mathcal{C}^{(g)}\cap \mathcal{S}^{-}(a^\ast)|
}{
|\mathcal{S}^{-}(a^\ast)|+\epsilon
}.
\end{equation}
Thus, rationales that mention KG-supported diagnostic relations receive positive reward, while rationales that invoke graph-incompatible relations are penalized. This provides a compact approximation of logical consistency under the closed-world FAKG schema.

By integrating GCRO, UniShield is encouraged to produce predictions with rationales that are more consistent with FAKG-supported forensic cues.

\noindent\textbf{Group-relative Advantage and Update.}
GCRO estimates advantages by normalizing rewards within each group (without a learned critic):
\begin{equation}
\hat{A}^{(g)}=\frac{R^{(g)}-\mu_R}{\sigma_R+\epsilon},
\quad
\mu_R=\frac{1}{G}\sum_{g=1}^{G}R^{(g)},
\quad
\sigma_R=\sqrt{\frac{1}{G}\sum_{g=1}^{G}\big(R^{(g)}-\mu_R\big)^2}.
\end{equation}
We then optimize the policy with a group-relative objective:
\begin{equation}
\mathcal{L}_{\text{GCRO}}
=
-
\mathbb{E}\left[\log \pi_{\theta}\big(y^{(g)} \mid x, q\big)\cdot \hat{A}^{(g)}\right],
\end{equation}
where $\hat{A}^{(g)}$ is treated as a constant w.r.t.\ $\theta$.
By combining AGIT and GCRO, the model is encouraged to improve both prediction accuracy and KG-consistent reasoning.

\section{Experimental Results}
\label{experiments}
\subsection{Experimental Settings}
\begin{figure}[t]
\centering

% ===================== 第一行：数据分布图 + 标签表 =====================

\begin{minipage}[t]{0.38\linewidth}
  \vspace{0pt}
  \centering
  \includegraphics[width=0.92\linewidth]{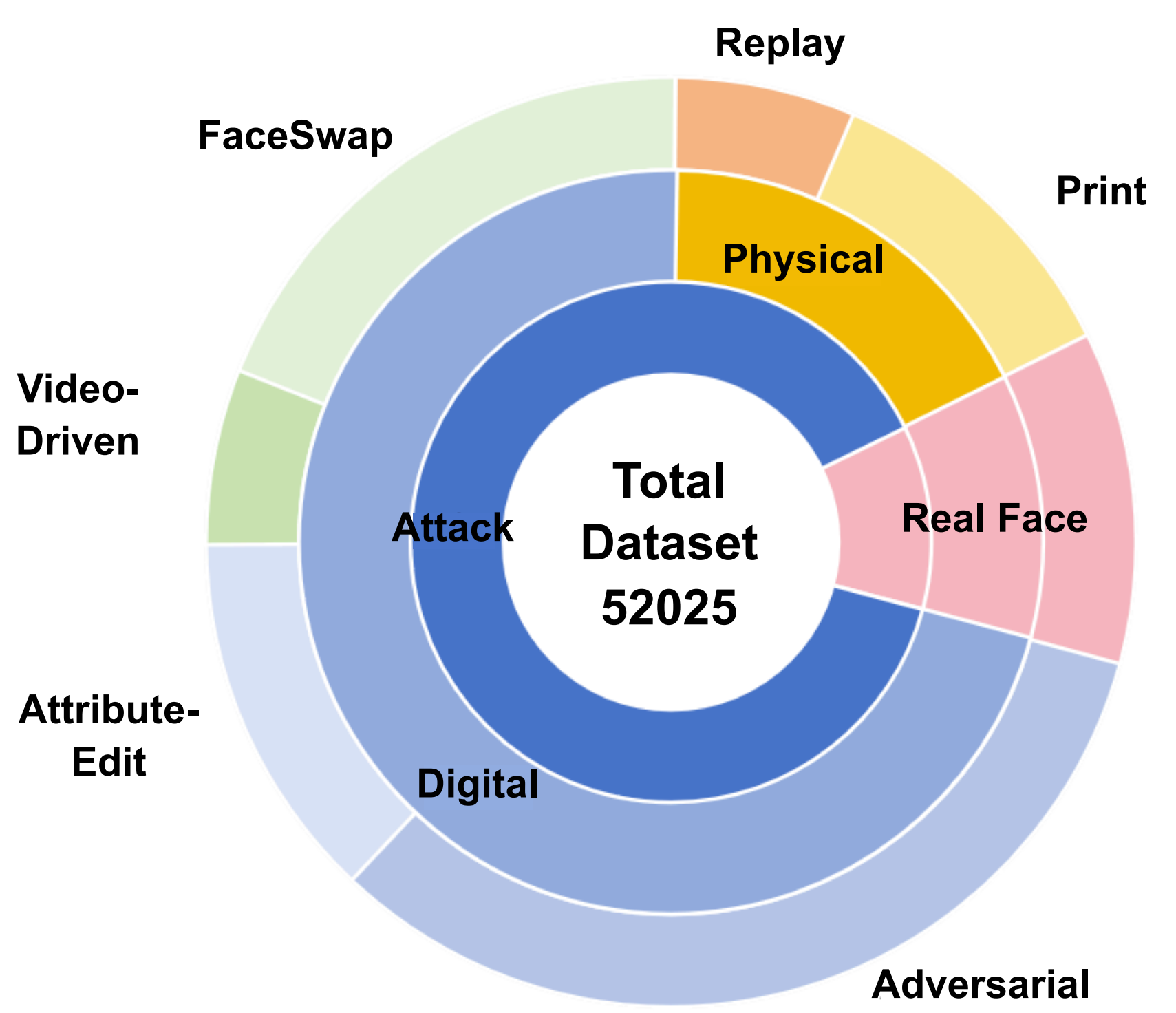}
  \captionof{figure}{FAKG-QA dataset composition. 
  Inner/middle/outer rings indicate binary, coarse, and fine-grained labels.}
  \label{fig:datapie}
\end{minipage}
\hfill
\begin{minipage}[t]{0.60\linewidth}
  \vspace{0pt}
  \centering

  \captionof{table}{Label definitions and data sizes under different protocols.}
  \label{tab:label_def}

  \vspace{-1mm}
  \scriptsize
  \setlength{\tabcolsep}{2.4pt}
  \renewcommand{\arraystretch}{0.82}

  \resizebox{0.92\linewidth}{!}{
  \begin{tabular}{c c l c c}
  \toprule
  Data & Prot. & Labels 
  & \makecell[c]{AGIT\\Size}
  & \makecell[c]{GCRO\\Size} \\
  \midrule

  Train 
  & -- 
  & \makecell[l]{Real Face, \{Pr, Re, FS,\\ AE, VD, Adv\}}
  & 41000
  & 4100 \\

  \midrule

  % \multirow{3}{*}{Test}
  % & Prot.1 
  % & Real Face, Attack
  % & \multirow{3}{*}{4706}
  % & \\

  % \cmidrule(lr){2-3}
  % & Prot.2 
  % & Real Face, Physical, Digital
  % & & \\

  % \cmidrule(lr){2-3}
  % & Prot.3 
  % & \makecell[l]{Real Face, \{Pr, Re, FS,\\ AE, VD, Adv\}}
  % & & \\
  
  \multirow{3}{*}{Test}
& Prot.1 
& Real Face, Attack
& \multicolumn{2}{c}{\multirow[c]{5}{*}{4706}} \\

\cmidrule(lr){2-3}
& Prot.2 
& Real Face, Physical, Digital
& \multicolumn{2}{c}{} \\

\cmidrule(lr){2-3}
& Prot.3 
& \makecell[l]{Real Face, \{Pr, Re, FS,\\ AE, VD, Adv\}}
& \multicolumn{2}{c}{} \\

  \bottomrule
  \end{tabular}
  }

  \vspace{2pt}
  \scriptsize
  \makecell[l]{
  \textit{Pr}: Print; \textit{Re}: Replay; \textit{FS}: FaceSwap;\\
  \textit{AE}: Attribute-Edit; \textit{VD}: Video-Driven; 
  \textit{Adv}: Adversarial.
  }

\end{minipage}

% ===================== 第二行：示例图 =====================

\vspace{3pt}

\includegraphics[width=0.96\linewidth]{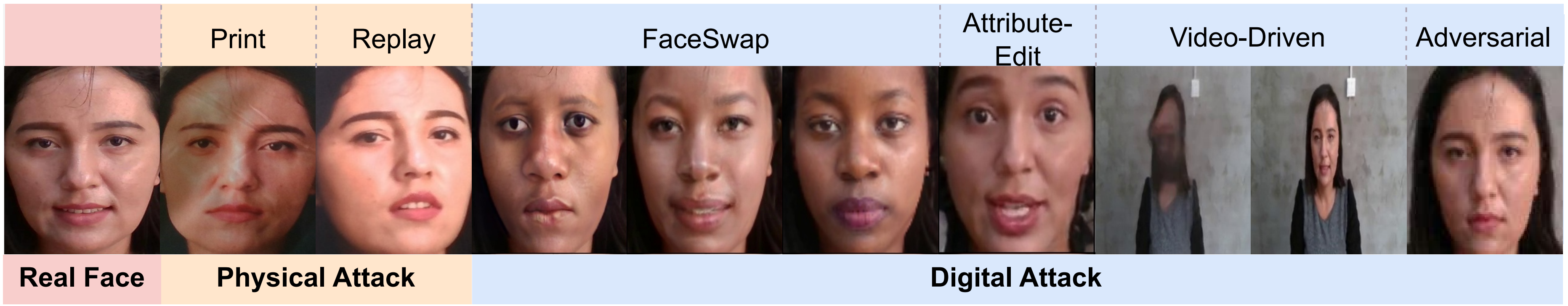}
\caption{Example images in our dataset.}
\label{fig:dataset_overview}
\vspace{-1mm}

\end{figure}

\noindent\textbf{Datasets.}
% 需要的宏包（导言区）
In this study, we build upon the UniAttackData dataset~\cite{uni}, which is currently the largest publicly available open-source benchmark for UAD, containing 28,706 videos (1,800 live and 26,906 fake) spanning 2 physical attack types and 12 digital attack types. On top of UniAttackData, we construct a customized FAKG-QA dataset with 52,025 high-quality QA pairs covering diverse attack categories, providing structured supervision for training and evaluating UAD models under heterogeneous spoofing scenarios. The FAKG-QA dataset integrates data from well-established datasets such as CASIA-SURF and CeFA~\cite{liucefa}, using these datasets as the basis for generating QA pairs and providing broad coverage of both physical and digital face spoofing attacks. The dataset composition and representative examples are shown in Fig.~\ref{fig:datapie} and Fig.~\ref{fig:dataset_overview}.
% UniAttackData stands out by offering a comprehensive range of both physical and digital attacks, addressing the incomplete coverage of attack types in existing datasets, making it a valuable benchmark for UAD methods.

\noindent\textbf{Implementation Details.}
We instantiate UniShield with Qwen2.5-VL~\cite{qwen} backbones, including both 3B and 7B variants, to evaluate the scalability of our framework.
All experiments are conducted on three NVIDIA A100 GPUs with 80GB memory.
Our training pipeline consists of two stages: Attack-Graph Instruction Tuning (AGIT) and Graph-Consistent Reasoning Optimization (GCRO).
In the AGIT stage, we perform full-parameter fine-tuning by updating the vision encoder, multimodal projector, and language model.
In the GCRO stage, implemented with GRPO, we freeze the vision encoder and multimodal projector for efficiency, and optimize only the language model.
As shown in our results, this two-stage training strategy consistently improves both 3B and 7B variants, with the final 7B variant achieving the best overall performance.

% \noindent\textbf{Protocols and Metrics.}
% As summarized in Tab.~\ref{tab:label_def}, we evaluate UAD methods under three testing protocols with increasing label granularity.
% During training, we use the full set of labels (i.e., all attack types) for all protocols; during testing, metrics are computed according to the protocol-specific label space:
% \vspace{-0.5em}
% \begin{itemize}\itemsep0.15em \topsep0.15em
%   \item \textbf{Prot.1 (Binary):} \textit{Real Face} vs.\ \textit{Attack}.
%   \item \textbf{Prot.2 (Coarse):} \textit{Real Face}, \textit{Physical}, \textit{Digital}.
%   \item \textbf{Prot.3 (Fine-grained):} All 7 attack types.
% \end{itemize}\vspace{-0.4em}

% We evaluate peformance using ACC and HTER, where HTER is the average of the false rejection rate (FRR) and the false acceptance rate (FAR).

\noindent\textbf{Protocols and Metrics.}
As summarized in Tab.~\ref{tab:label_def}, all models are trained with the full fine-grained label space and evaluated under three protocol-specific label spaces:
\vspace{-0.5em}
\begin{itemize}\itemsep0.15em \topsep0.15em
  \item \textbf{Prot.1 (Binary):} \textit{Real Face} vs.\ \textit{Attack}.
  \item \textbf{Prot.2 (Coarse):} \textit{Real Face}, \textit{Physical}, \textit{Digital}.
  \item \textbf{Prot.3 (Fine-grained):} \textit{Real Face} and six fine-grained attack types.
\end{itemize}\vspace{-0.4em}

We report accuracy (ACC) and half total error rate (HTER).
For Prot.1, HTER follows the standard binary definition, i.e.,
\begin{equation}
\mathrm{HTER}=\frac{1}{2}(\mathrm{FAR}+\mathrm{FRR}).
\end{equation}
For Prot.2/3, HTER is computed over protocol-specific categories with a one-vs-rest formulation.
Thus, cross-attack misclassification is counted as a category-level error in the corresponding multi-class protocol.
For the \#Total row, ACC and HTER are computed as weighted averages of the corresponding category-level metrics over protocol-specific categories, with each category weighted by its number of images.

\noindent\textbf{Compared Methods.}
We train and evaluate UniShield on our proposed dataset against five representative baselines, including CDCN~\cite{cdcn}, CLIP~\cite{clip}, CoOp~\cite{zhou2022coop}, UniAttackDetection~\cite{uni}, and HiPTune~\cite{uni++}.
For VLM-based methods, we follow the data preprocessing in UniAttackDetection~\cite{uni} to ensure a fair comparison.
Since existing face-security MLLM methods are mainly designed for single face anti-spoofing or forgery detection tasks rather than unified physical--digital attack detection, we instead select strong general-purpose MLLMs for evaluation.
In addition, we compare UniShield with state-of-the-art MLLMs, including GLM-4.6V~\cite{glm}, GPT-5.2~\cite{gpt5.2}, and Gemini-3-Pro~\cite{gemini3pro}.

\subsection{Comparison Results}

\noindent\textbf{Compared to SOTA methods.}
Table~\ref{tab:prot12_merged} and Table~\ref{tab:full_breakdown} compare UniShield with representative SOTA methods under three protocols.
UniShield consistently achieves the best overall performance across all settings.
In Prot.1, UniShield (7B) reaches 99.6\% ACC and 0.47\% HTER, surpassing prior specialized methods such as UniAttackDetection and HiPTune.
In Prot.2, it further obtains the best overall result with 99.5\% ACC and 0.36\% HTER on \#Total, while maintaining balanced performance across real-face, physical, and digital categories.
More importantly, under the fine-grained Prot.3 setting, UniShield (7B) achieves 99.4\% ACC and 0.38\% HTER.
The consistent gains across diverse attack types, including both physical and digital attacks, demonstrate that attack-graph aligned reasoning improves not only binary spoof detection but also fine-grained attack discrimination.

\noindent\textbf{Compared to SOTA MLLMs.}
We further evaluate strong general-purpose MLLMs, including GLM-4.6V, GPT-5.2, and Gemini-3-Pro.
Despite their general visual-language capabilities, these models perform poorly under the proposed protocols.
For example, the best-performing MLLM on Prot.1, GPT-5.2, achieves only 80.2\% ACC and 42.4\% HTER, far behind UniShield (7B).
The gap becomes more pronounced in Prot.3, where all evaluated MLLMs obtain below 30\% overall ACC.
These results indicate that off-the-shelf MLLMs lack the domain-specific grounding and decision calibration required for unified attack detection.
In contrast, UniShield explicitly aligns visual observations with structured attack knowledge, enabling more accurate and reliable predictions across protocols and attack categories.

\begin{figure}[H]
  \centering
  \includegraphics[width=0.95\linewidth]{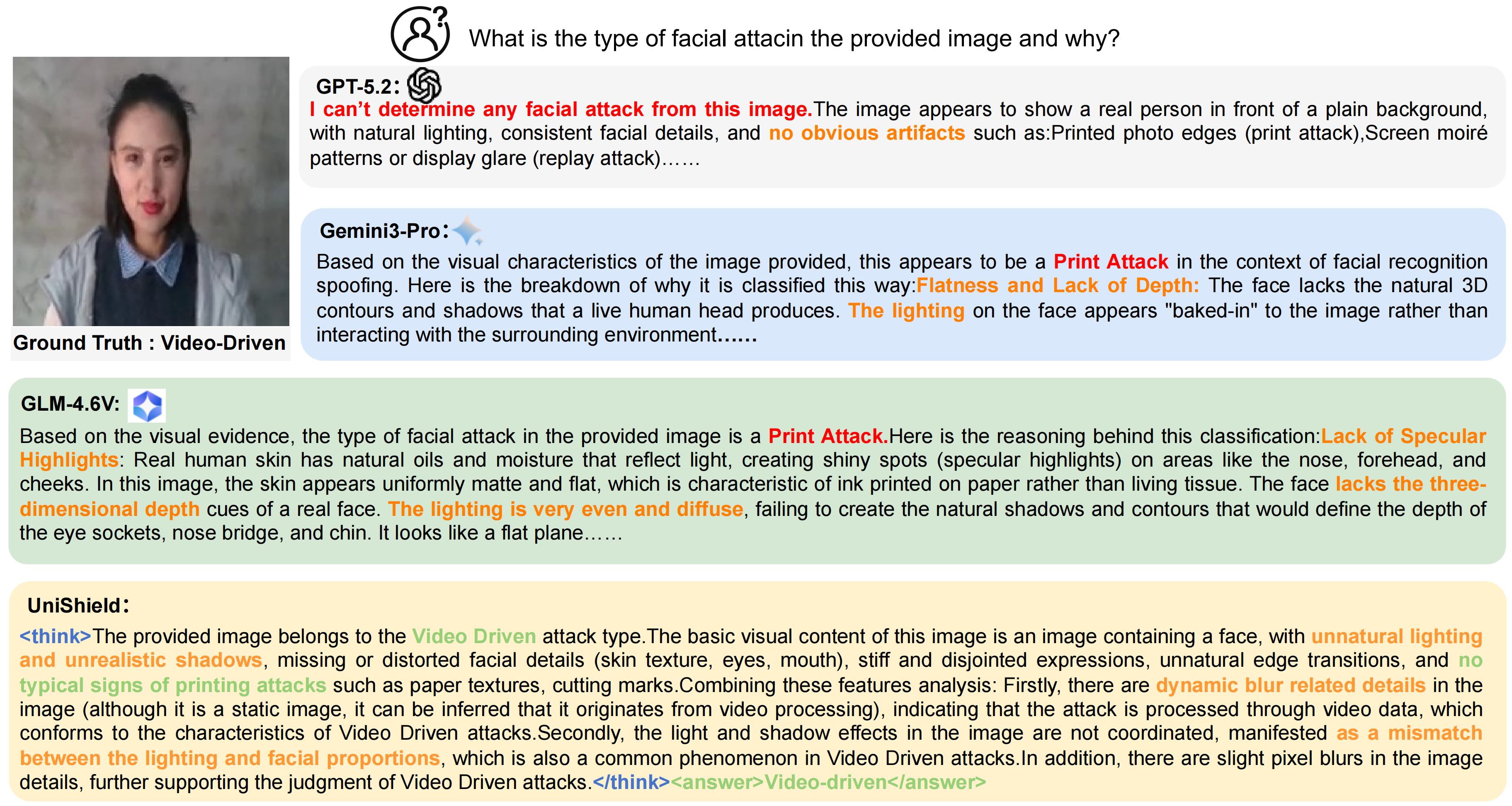}
  \caption{Comparison of multimodal models on the UAD task}
  \label{fig:QA answer}
% \vspace{-5mm}
\end{figure}

Beyond quantitative gaps, Fig.~\ref{fig:QA answer} provides a representative qualitative case: given a Video-Driven attack (ground truth), GPT-5.2 fails to make a decision, while Gemini-3-Pro and Qwen3.5-Plus incorrectly predict a Print attack by relying on superficial cues (e.g., perceived flatness or lighting patterns).
In contrast, UniShield correctly identifies the sample as Video-Driven and, guided by a more reliable reasoning process, provides the correct prediction with consistent supporting evidence.
% In this section, we present a comprehensive comparison between traditional methods, MLLMs, and our proposed approach, tested across the three protocols. The results demonstrate that our method outperforms the traditional approaches and other MLLMs in terms of accuracy (ACC) and half-total error rate (HTER). Specifically, our method achieves good performance on all three protocols, as shown in Tables 3 and Table 4.
\begin{table*}[t]
\centering
\caption{Performance comparison under Protocol 1 and Protocol 2.}
\label{tab:prot12_merged}

\setlength{\tabcolsep}{3pt}
\renewcommand{\arraystretch}{1.08}
\scriptsize

\resizebox{\textwidth}{!}{%
\begin{tabular}{lcccccccccc}
\toprule
\multirow{3}{*}{Method}
& \multicolumn{2}{c}{Prot.1}
& \multicolumn{8}{c}{Prot.2} \\
\cmidrule(lr){2-3}\cmidrule(lr){4-11}
& \multirow{2}{*}{ACC} & \multirow{2}{*}{HTER}
& \multicolumn{2}{c}{Real Face}
& \multicolumn{2}{c}{Physical}
& \multicolumn{2}{c}{Digital}
& \multicolumn{2}{c}{\#Total} \\
\cmidrule(lr){4-5}\cmidrule(lr){6-7}\cmidrule(lr){8-9}\cmidrule(lr){10-11}
&  &  & ACC & HTER & ACC & HTER & ACC & HTER & ACC & HTER \\
\midrule

CDCN~\cite{cdcn} & 98.4 & 1.17 & 89.0 & 11.0 & 97.3 & 2.67  & 98.9 & 1.11  & 98.0 & 2.04 \\
CLIP~\cite{clip} & 93.6 & 50.0 & 86.7 & 6.67  & 99.1  & 0.44  & \textbf{99.6} & \underline{0.23}  & 98.7 & 0.68 \\
CoOp~\cite{zhou2022coop} & 92.8 & 8.82  & 75.7 & 12.2 & \underline{99.7}  & \textbf{0.17}  & 98.4 & 0.79  & 97.2 & 1.39 \\
UniAttackDetection~\cite{uni} & \underline{99.0} & \underline{0.97} & 88.0 & 6.00  & 99.1  & 0.44  & \textbf{99.6} & \textbf{0.19}  & 98.8 & \underline{0.61} \\
HiPTune~\cite{uni++} & 98.4 & 1.13 & 96.1 & 1.96 & 99.3 & 0.33 & 98.7 & 0.67 & \underline{98.9} & 0.69 \\

\specialrule{0.03em}{2pt}{2pt}

GLM-4.6V~\cite{glm} & 8.50  & 49.7 & \textbf{100}& \textbf{0.00}  & 0.00   & 100 & 0.43  & 49.8 & 6.70  & 56.2 \\
GPT-5.2~\cite{gpt5.2} & 80.2 & 42.4 & 38.3 & 30.8 & 1.67   & 49.2  & 67.0 & 27.0 & 53.0 & 31.5 \\
Gemini-3-Pro~\cite{gemini3pro} & 31.4 & 36.7 & \textbf{100}& \textbf{0.00}  & 72.8  & 26.1  & 14.6 & 42.9 & 31.4 & 36.9 \\

\specialrule{0.03em}{2pt}{2pt}

\rowcolor{rowgray}
UniShield (3B) & 98.6 & 1.42 & 99.0 & 1.00 & 97.2 & 1.40 & 98.9 & 1.09 & 98.6 & 1.14 \\

\rowcolor{rowgray}
UniShield (7B)  & \textbf{99.6} & \textbf{0.47} & \underline{99.3} & \underline{0.67}  & \textbf{99.8}  & \underline{0.22}  & \underline{99.4} & 0.37  & \textbf{99.5} & \textbf{0.36} \\

\bottomrule
\end{tabular}
}
\end{table*}
\vspace{-2mm}

% In particular, under Prot.~1, our UniShield model delivers the strongest overall performance, reaching 99.25\% ACC with only 0.70\% HTER, indicating highly reliable attack discrimination even in the basic evaluation setting. Moving to the more fine-grained Prot.~2, our approach remains consistently robust across different subsets (Live / Physical / Digital) and attains 99.22\% ACC and 0.78\% HTER on Total, showing that the learned representations generalize well to diverse attack sources rather than overfitting to a single subset. Notably, while several classical baselines perform competitively on specific categories, their overall performance degrades when evaluated across mixed conditions, revealing limited cross-type generalization.

\begin{table*}[t]
\centering
\scriptsize
\setlength{\tabcolsep}{3pt}
\renewcommand{\arraystretch}{1.20}
\caption{Performance comparison under Protocol 3.}

\label{tab:full_breakdown}

\resizebox{\textwidth}{!}{
\begin{tabular}{lcccccccccccccccc}
\toprule
\multirow{2}{*}{Method}
& \multicolumn{2}{c}{Real Face}
& \multicolumn{2}{c}{Print}
& \multicolumn{2}{c}{Replay}
& \multicolumn{2}{c}{FaceSwap}
& \multicolumn{2}{c}{Attribute-Edit}
& \multicolumn{2}{c}{Video-Driven}
& \multicolumn{2}{c}{Adversarial}
& \multicolumn{2}{c}{\#Total} \\

\cmidrule(lr){2-3}
\cmidrule(lr){4-5}
\cmidrule(lr){6-7}
\cmidrule(lr){8-9}
\cmidrule(lr){10-11}
\cmidrule(lr){12-13}
\cmidrule(lr){14-15}
\cmidrule(lr){16-17}

& ACC & HTER & ACC & HTER & ACC & HTER
& ACC & HTER & ACC & HTER
& ACC & HTER & ACC & HTER
& ACC & HTER \\

\midrule

CDCN~\cite{cdcn}
& 1.33 & 49.3
& 76.2 & 11.9
& 66.3 & 16.8
& 38.5 & 30.7
& 25.3 & 37.3
& 16.5 & 41.8
& 3.28 & 48.4
& 26.3 & 36.9 \\

CLIP~\cite{clip}
& 25.7 & 37.2
& 99.3 & \underline{0.33}
& \underline{99.3} & \underline{0.33}
& \textbf{98.9} & \textbf{0.56}
& \underline{98.7} & \underline{0.67}
& 99.0 & 0.50
& \textbf{100} & \textbf{0.00}
& 94.7 & 2.67 \\

CoOp~\cite{zhou2022coop}
& 78.0 & 11.0
& 99.3 & \underline{0.33}
& 99.0 & 0.50
& 96.1 & 1.96
& \textbf{99.7} & \textbf{0.17}
& 93.8 & 3.08
& \textbf{100} & \textbf{0.00}
& 96.9 & 1.55 \\

UniAttackDetection~\cite{uni}
& 85.3 & 7.33
& 99.3 & \underline{0.33}
& 98.7 & 0.67
& \textbf{98.9} & \textbf{0.56}
& \underline{98.7} & \underline{0.67}
& 98.2 & 0.92
& \textbf{100} & \textbf{0.00}
& 98.4 & 0.82 \\

HiPTune~\cite{uni++}
& 97.3 & 2.67
& \textbf{99.8} & \textbf{0.17}
& \textbf{100} & \textbf{0.00}
& 88.1 & 11.9
& 98.0 & 1.00
& 99.0 & 0.50
& \underline{99.9} & \underline{0.06}
& 97.3 & 2.58 \\

\specialrule{0.01em}{1pt}{2pt}

GLM-4.6V~\cite{glm}
& \textbf{100} & \textbf{0.00}
& 15.0 & 85.0
& 1.67 & 93.3
& 0.00 & 100
& 0.00 & 100
& 0.00 & 100
& 0.00 & 100
& 8.40 & 91.3 \\

GPT-5.2~\cite{gpt5.2}
& 61.7 & 38.3
& 38.3 & 45.0
& 2.00 & 53.3
& 0.00 & 100
& 0.00 & 100
& 0.00 & 100
& 0.29 & 49.9
& 9.05 & 68.0 \\

Gemini-3-Pro~\cite{gemini3pro}
& 95.0 & 5.00
& 60.0 & 20.0
& 66.7 & 18.3
& 19.0 & 40.5
& 0.00 & 100
& 13.3 & 43.4
& 17.3 & 41.4
& 29.5 & 39.0 \\

\specialrule{0.03em}{1pt}{2pt}

\rowcolor{rowgray}
UniShield (3B)
& 99.0 & 1.00
& 98.8 & 0.83
& 94.0 & 3.00
& 97.3 & 1.34
& 96.7 & 1.67
& \underline{99.3} & \underline{0.33}
& \underline{99.9} & \underline{0.06}
& \underline{98.6} & \underline{0.79} \\

\rowcolor{rowgray}
UniShield (7B)
& \underline{99.3} & \underline{0.67}
& \underline{99.7} & \underline{0.33}
& \underline{99.3} & \underline{0.33}
& \underline{98.4} & \underline{1.00}
& 97.3 & 1.33
& \textbf{100} & \textbf{0.00}
& \textbf{100} & \textbf{0.00}
& \textbf{99.4} & \textbf{0.38} \\

\bottomrule
\end{tabular}
}
\vspace{-1mm}
\end{table*}

\subsection{Ablation Study}
\begin{figure}[H]
\centering
\small

\begin{minipage}[t]{0.56\linewidth}
\centering
\vspace{0pt}
\setlength{\tabcolsep}{3.2pt}
\renewcommand{\arraystretch}{1.12}

\captionof{table}{Ablation results under three protocols. Gray rows mark key methods; colored values show changes from the previous row.}
\label{tab:ablation_all}

\resizebox{\linewidth}{!}{
\begin{tabular}{
@{}l
!{\color{black!55}\vrule width 0.7pt}
cc
!{\color{black!55}\vrule width 0.7pt}
cc
!{\color{black!55}\vrule width 0.7pt}
cc@{}
}
\toprule
\multirow{2}{*}{Method}
& \multicolumn{2}{c}{Prot.1}
& \multicolumn{2}{c}{Prot.2}
& \multicolumn{2}{c}{Prot. 3} \\
\cmidrule(lr){2-3}
\cmidrule(lr){4-5}
\cmidrule(lr){6-7}
& ACC & HTER
& ACC & HTER
& ACC & HTER \\
\midrule

\rowcolor{rowgray}
\textbf{UniShield-3B Backbone}
& 16.3 & 40.4
& 10.3 & 48.7
& 14.9 & 39.9 \\

\quad + CoT
& 83.4 & 13.7
& 83.2 & 4.98
& 83.4 & 4.98 \\

\rowcolor{rowgray}
\quad + AGIT
& 92.2 {\gup{8.8}}
& 4.18 {\gdown{9.52}}
& 92.1 {\gup{8.9}}
& 4.80 {\gdown{0.18}}
& 94.0 {\gup{10.6}}
& 3.25 {\gdown{1.73}} \\

\quad + AGIT+GRPO
& 97.9 & 2.58
& 98.4 & 2.04
& 98.4 & 1.19 \\

\rowcolor{rowgray}
\quad + AGIT+GCRO
& \textbf{98.6} {\gup{0.7}}
& \textbf{1.42} {\gdown{1.16}}
& \textbf{98.6} {\gup{0.2}}
& \textbf{1.14} {\gdown{0.90}}
& \textbf{98.6} {\gup{0.2}}
& \textbf{0.79} {\gdown{0.40}} \\

\midrule

\rowcolor{rowgray}
\textbf{UniShield-7B Backbone}
& 7.86 & 48.6
& 13.3 & 45.7
& 8.7 & 45.6 \\

  \quad + AGIT
  & 99.3 & 0.67
  & 99.2 & 0.78
  & 98.8 & 0.71 \\

  \rowcolor{rowgray}
  \quad + AGIT+GCRO
  & \textbf{99.6} {\gup{0.3}}
  & \textbf{0.47} {\gdown{0.20}}
  & \textbf{99.5} {\gup{0.3}}
  & \textbf{0.36} {\gdown{0.42}}
  & \textbf{99.4} {\gup{0.6}}
  & \textbf{0.38} {\gdown{0.33}} \\

\bottomrule
\end{tabular}
}
\end{minipage}
\hfill
\begin{minipage}[t]{0.41\linewidth}
\centering
\vspace{0pt}

\includegraphics[width=\linewidth]{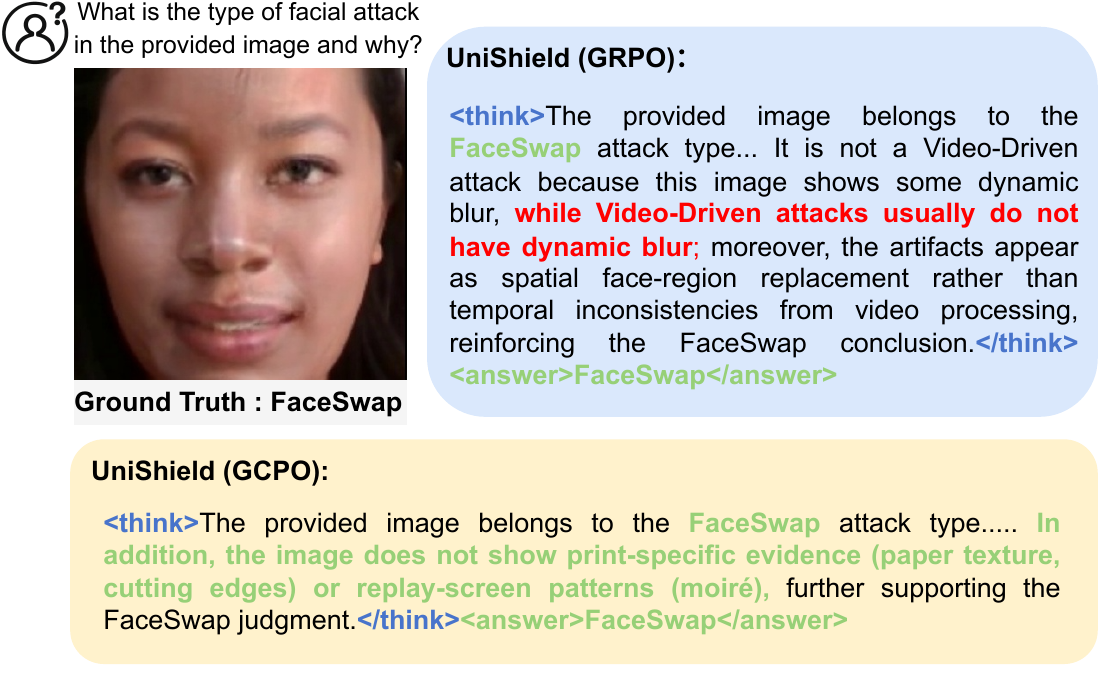}

\captionof{figure}{Performance comparison between GRPO and GCRO.}
\label{fig:qagcro}
\end{minipage}

\vspace{-1mm}
\end{figure}

We conduct ablation studies to quantify the contributions of our two key components, AGIT and GCRO. As shown in Tab.~\ref{tab:ablation_all}, AGIT substantially improves over model trained by standard CoT dataset on the 3B model, increasing ACC by 8.8\%, 8.9\%, and 10.6\% and reducing HTER by 9.52\%, 0.18\%, and 1.73\% on Prot.~1--3, respectively. This demonstrates that attack-graph-guided instruction tuning provides more effective supervision than generic chain-of-thought reasoning. Starting from the AGIT-tuned model, GCRO further outperforms vanilla GRPO by incorporating graph-consistency rewards, bringing additional ACC gains of 0.7\%, 0.2\%, and 0.2\% and HTER reductions of 1.16\%, 0.90\%, and 0.40\% on Prot.~1--3. The 7B results show the same trend, where applying GCRO after AGIT improves ACC/HTER from 99.3\%/0.67\%, 99.2\%/0.78\%, and 98.8\%/0.71\% to 99.6\%/0.47\%, 99.5\%/0.36\%, and 99.4\%/0.38\%. Fig.~\ref{fig:qagcro} further shows that GCRO produces rationales more aligned with FAKG-supported attack cues than GRPO.

%

%\paragraph{Effect of Our Dataset.}
% To isolate the contribution of our dataset design, we compare SFT under three data conditions: (i) SFT on our KG-guided QA dataset (AGIT), (ii) SFT on QA pairs constructed without KG grounding (w/o AGIT), and (iii) the corresponding model trained with our full KG-guided setting (UniShield). The comparison shows that the performance gains mainly come from the KG-aware data construction itself: when KG grounding is removed during dataset construction, the SFT model exhibits a substantial degradation across all protocols, indicating that generic SFT on ungrounded QA is insufficient. In contrast, AGIT provides structured, attack-conditioned feature cues that guide the model to learn more reliable evidence and improves generalization consistently under Prot.~1--3, yielding results close to the full UniShield configuration (Table~4).

\vspace{-1mm}
\section{Conclusion}
\label{conclusion}
% This paper studied unified face attack detection, where a single model must handle heterogeneous physical and digital attacks with consistent semantics. We proposed a knowledge-graph-driven MLLM framework that (i) constructs a unified face attack knowledge graph linking attack types to observable cues and underlying causes, (ii) uses the graph to generate knowledge-driven QA data for supervised fine-tuning, and (iii) introduces a knowledge-guided reinforcement learning strategy to regularize reasoning trajectories toward graph-consistent explanations. We also release a new benchmark for unified face attack detection. Comprehensive experiments conducted on our benchmark and existing protocols show that our method outperforms current state-of-the-art approaches, delivering more reliable and interpretable results.

In this paper, we presented UniShield, a knowledge-grounded multimodal reasoning framework for unified face attack detection. By introducing the FAKG, we organize attack categories and diagnostic cues into a structured semantic space across heterogeneous physical and digital attack modalities. Our approach leverages AGIT and GCRO to encourage rationales that are more consistent with graph-supported forensic cues. Extensive evaluations on our multimodal UAD benchmark show that UniShield achieves strong detection performance while providing more interpretable predictions. Nevertheless, our evaluation is mainly conducted on a UniAttackData-derived benchmark, and FAKG currently relies on predefined attack-cue relations. Extending the framework to independent third-party benchmarks and adapting it to unseen attack types remain important future directions.

% ---- Bibliography ----
\bibliographystyle{plainnat}
\bibliography{main}

\appendix
\section{FAKG-QA Construction and Detailed Prompting}

\subsection{FAKG Knowledge Graph Construction}
To construct the Face Attack Knowledge Graph (FAKG), we leverage the UniAttackData dataset to define key attack types, including Print, Replay, FaceSwap, Video-Driven, Adversarial, and Attribute-Edit, each representing a distinct spoofing pattern. These types form the primary semantic nodes in the graph, as detailed in Algorithm~\ref{alg:fakg}.

Each attack type is linked to common visual and physical features—such as lighting inconsistencies, abnormal textures, and boundary artifacts—that help differentiate live faces from spoofed samples.

Relationships between attack types and features are initially based on statistical data from the dataset and further refined using insights from face anti-spoofing literature. MLLMs are utilized to strengthen semantic connections by generating structured reasoning paths that map visual cues to attack mechanisms.

Expert knowledge is incorporated to refine and validate feature–attack relationships, particularly for attack types with overlapping visual characteristics (e.g., replay and video-driven). This hybrid approach ensures the FAKG is both structurally sound and semantically rich, providing a robust foundation for downstream reasoning and model training.

\begin{algorithm}[h]
\caption{Construction of the Face Attack Knowledge Graph (FAKG)}
\label{alg:fakg}
\begin{algorithmic}[1]
\REQUIRE UniAttackData dataset $D$, literature knowledge $L$, MLLM $M$, expert knowledge $E$
\ENSURE Face Attack Knowledge Graph $G$

\STATE Initialize graph $G=(V,R)$

\STATE Extract attack types $T$ from dataset $D$
\FOR{each attack type $t \in T$}
    \STATE Add node $t$ into $V$
    \STATE Extract feature set $F_t$ from dataset statistics
    \FOR{each feature $f \in F_t$}
        \STATE Add node $f$ into $V$
        \STATE Add relation $(t,f)$ into $R$
    \ENDFOR
\ENDFOR

\FOR{each attack type $t \in T$}
    \STATE Use MLLM $M$ to generate additional feature candidates $F'_t$
    \FOR{each feature $f' \in F'_t$}
        \STATE Add node $f'$ into $V$ if not existing
        \STATE Add relation $(t,f')$ into $R$
    \ENDFOR
\ENDFOR

\STATE Extract feature relations $R_L$ from literature $L$
\FOR{each relation $(t,f) \in R_L$}
    \STATE Add node $f$ if not existing
    \STATE Add relation $(t,f)$ into $R$
\ENDFOR

\FOR{each relation $(t,f) \in R$}
    \IF{expert $E$ marks the relation as inconsistent}
        \STATE Modify or remove $(t,f)$
    \ENDIF
\ENDFOR

\RETURN Final knowledge graph $G$
\end{algorithmic}
\end{algorithm}

Once the FAKG is constructed, it is integrated into the framework to generate corresponding QA pairs. Each pair is designed to explain the attack type and its associated features. The QA pairs are derived from the knowledge graph, helping each prompt capture relevant attack-specific features and their relationships.

\subsection{FAKG-QA Detailed Prompt Design}
\begin{figure}[h]
  \centering
  \includegraphics[width=0.8\linewidth]{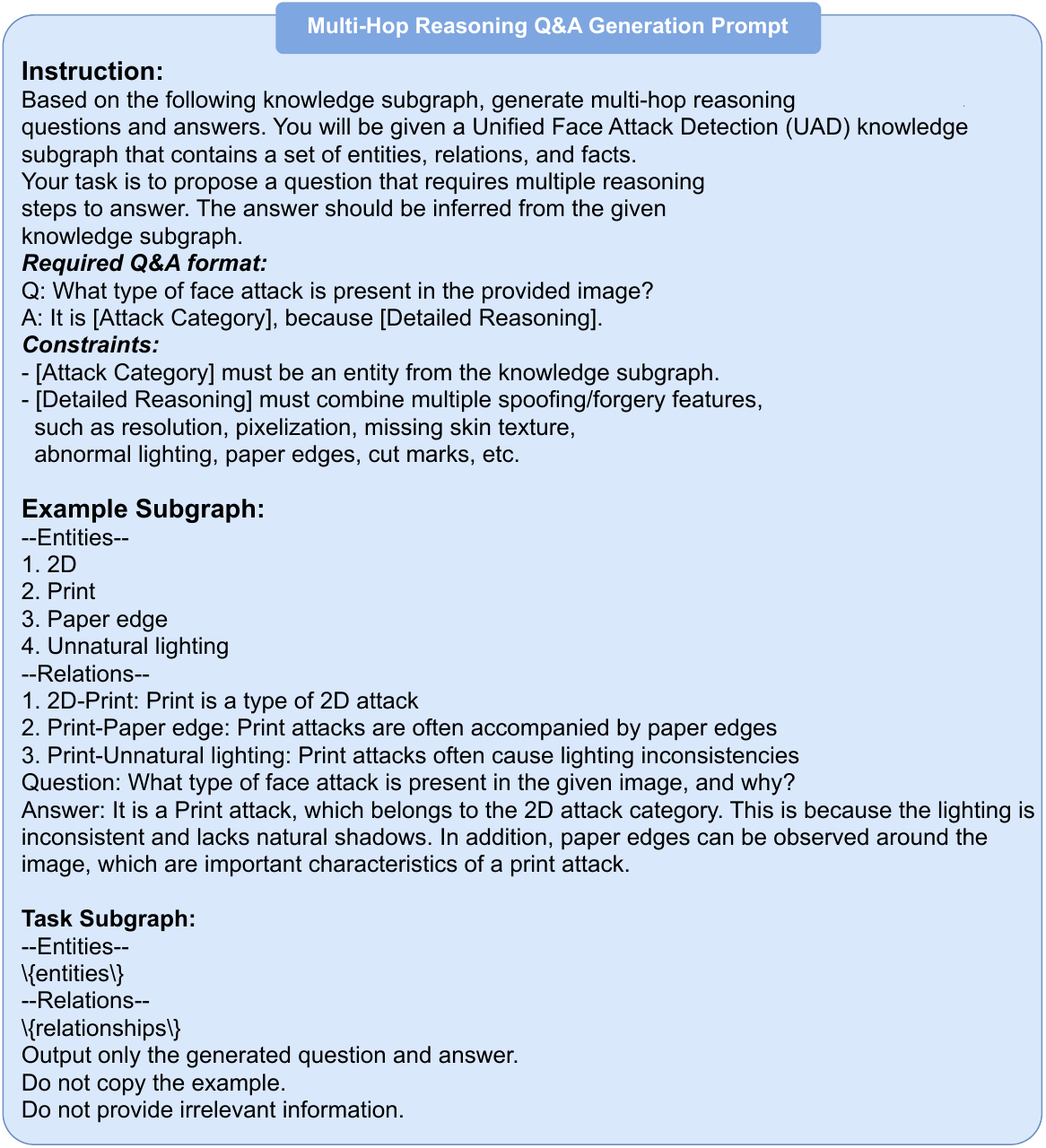}
  \caption{Multi-Hop Reasoning Q\&A Generation Prompt}
   \label{fig:mul_pr}
\end{figure}

We designed multiple prompts for generating the FAKG-QA pairs from the knowledge graph. These prompts allow the model to reason across different attack types and semantic levels. The prompts aim to extract both coarse-grained and fine-grained attack semantics. Here are the detailed prompts used:

\textbf{Prompt 1: Multi-Hop Reasoning Q\&A Generation.}
This prompt instructs GraphGen to construct a QA framework that generates coherent questions and answers based on the extracted attack types and relationships, leveraging the Qwen series of models. As shown in Fig.~\ref{fig:mul_pr}, the framework utilizes Qwen's capabilities to ensure accurate semantic alignment and context-aware generation.

\begin{figure}[h]
  \centering
  \includegraphics[width=0.8\linewidth]{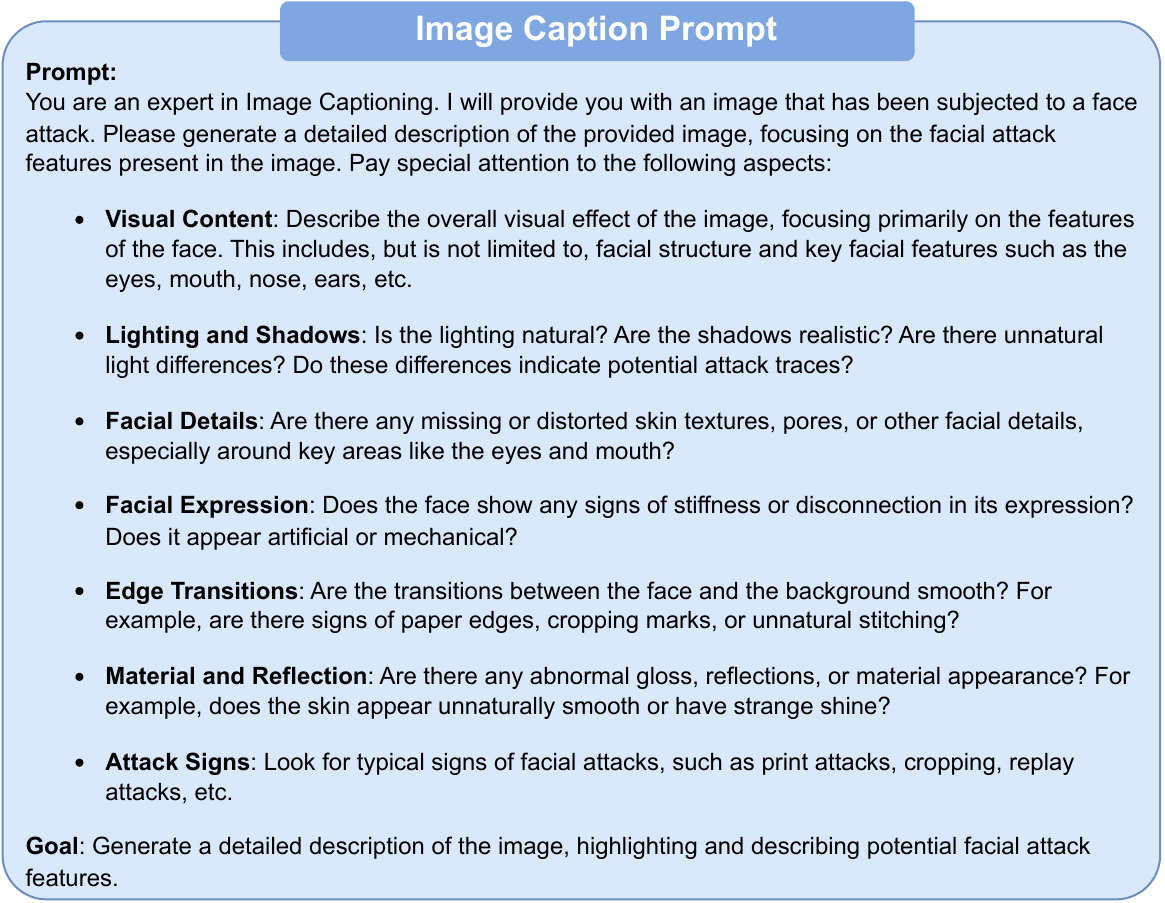}
  \caption{Image caption prompt}
    \label{fig:caption_pr}
\end{figure}

\textbf{Prompt 2: Generating Image Descriptions.}
This prompt directs GLM-4.1V to generate image captions, focusing on attack-specific visual features present in the provided image. As shown in Fig.~\ref{fig:caption_pr}, the model leverages its capabilities to ensure accurate identification and description of attack-related visual cues.

\begin{figure}[h]
  \centering
  \includegraphics[width=0.8\linewidth]{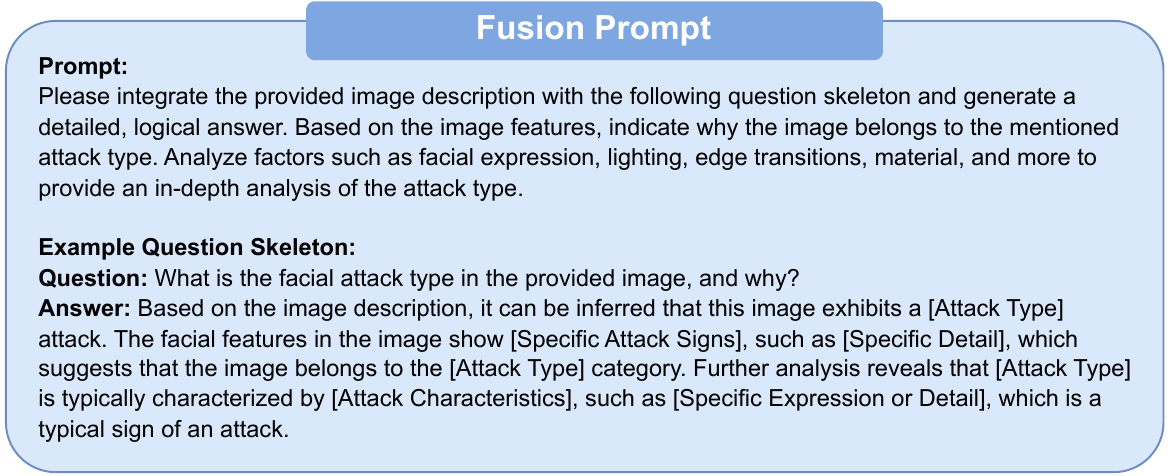}
  \caption{Fusion prompt}
    \label{fig:fusion_pr}
\end{figure}

\textbf{Prompt 3: Integrating the QA Framework and Image Descriptions.}
This prompt guides GLM-4.1V to fuse the generated QA framework with the image captions, producing a detailed and interpretable answer that links the visual cues to attack types. As shown in Fig.~\ref{fig:fusion_pr}, the fusion process ensures a coherent connection between the visual information and the semantic content of the attack.

\subsection{Data Purification and Consistency-Driven Filtering}

To ensure that the synthesized training samples maintain a high degree of causal consistency between expert knowledge and visual content, we design a multi-stage pre-fusion filtering pipeline. This process occurs after the generation of the QA reasoning framework but before the final multimodal fusion, acting as a rigorous quality gate.

\textbf{Semantic Consistency Verification via FAKG.}
Upon obtaining the QA reasoning framework from GraphGen, we first perform a structural validation. By parsing the entity-relation triplets $(a, f, \phi)$ within the framework, we match them against the canonical topology of the FAKG. Any combination that violates domain common sense (e.g., attributing "paper edge" artifacts to a "digital adversarial attack") is flagged as structurally invalid and discarded. This ensures the reasoning chain remains anchored to forensic ground truth at the knowledge level.

\textbf{Rationale Logical Flow Validation.}
Before the final fusion, we employ a MLLM as an intelligent verifier to identify deep semantic contradictions. An example prompt template is shown in Fig.~\ref{fig:lo_pr}.
\begin{itemize}
\item \textbf{Verification Logic:} The MLLM is prompted as a "Critic" to scrutinize the alignment between the image captions (describing raw visual facts) and the proposed QA framework (describing the attack logic). For instance, if the image caption identifies "pristine high-resolution textures" while the QA framework assumes "screen replaying" (which necessitates moir\'e patterns), the verifier will detect a factual mismatch.
\item \textbf{Execution Policy:} Samples marked with "Fact-Conflict" by the verifier are pruned. This mechanism effectively suppresses hallucination coupling, a common pitfall where incorrect logic is forcefully paired with mismatched visual evidence.

\begin{figure}[h]
  \centering
  \includegraphics[width=0.8\linewidth]{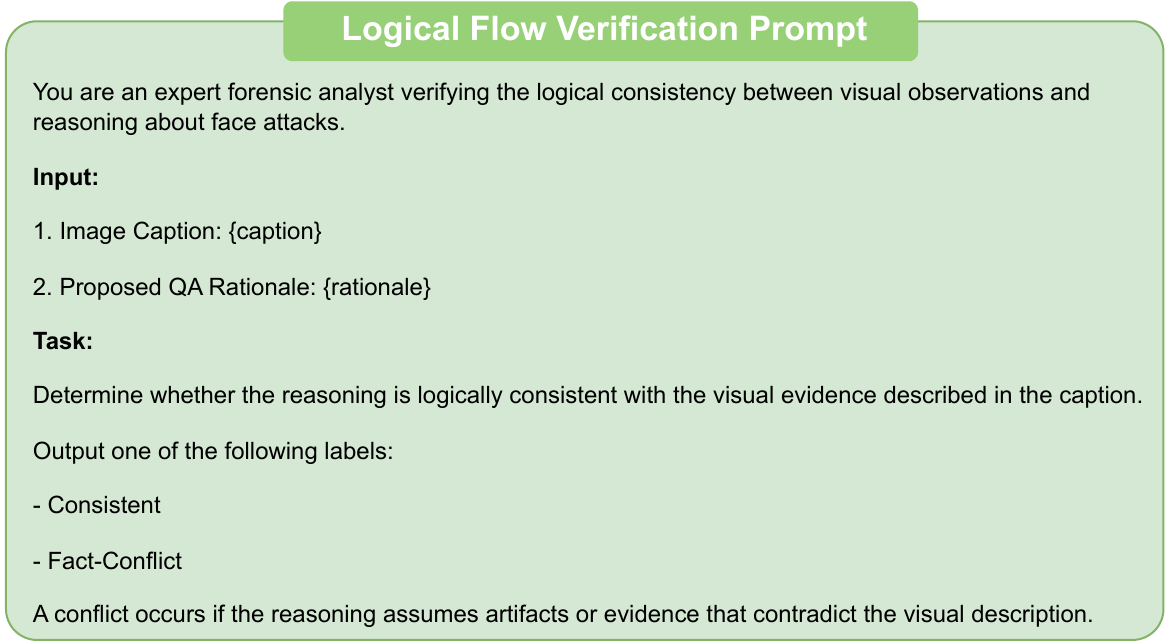}
  \caption{Prompt for Logical Flow Verification.}
    \label{fig:lo_pr}
\end{figure}

\end{itemize}
\subsubsection{Multi-Stage Pruning for Hallucination Mitigation.}
\begin{figure}[h]
  \centering
  \includegraphics[width=0.8\linewidth]{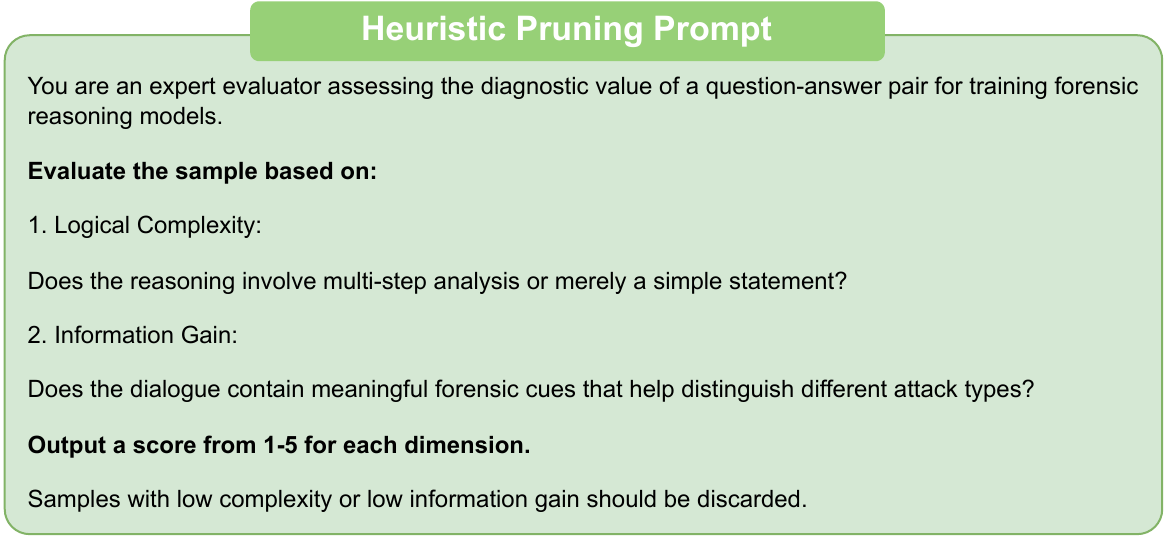}
  \caption{Prompt for Heuristic Pruning}
    \label{fig:pru_pr}
\end{figure}

We further leverage the MLLM's semantic understanding to perform heuristic pruning. An example prompt template is shown in Fig.~\ref{fig:pru_pr}:
\begin{itemize}
\item \textbf{Logical Complexity Scoring:} We prioritize samples featuring multi-hop reasoning and dense diagnostic evidence, filtering out overly simplistic or repetitive templates.
\item \textbf{Information Gain Filtering:} Generic dialogues with low diagnostic value are removed to ensure the model learns complex, "clinical-style" forensic logic during SFT.
\end{itemize}

\end{document}